# Player Pressure Map - A Novel Representation of Pressure in Soccer for Evaluating Player Performance in Different Game Contexts

Chaoyi Gu*, Jiaming Na*, Yisheng Pei*, Varuna De Silva

*Equal Contribution.

Institute: Institute for Digital Technologies, Loughborough University, London, United Kingdom

Contact: Chaoyi Gu, email: c.gu@lboro.ac.uk

## Abstract

In soccer, contextual player performance metrics are invaluable to coaches. For example, the ability to perform under pressure during matches distinguishes the elite from the average. Appropriate pressure metric enables teams to assess players' performance accurately under pressure and design targeted training scenarios to address their weaknesses. The primary objective of this paper is to leverage both tracking and event data and game footage to capture the pressure experienced by the possession team in a soccer game scene.

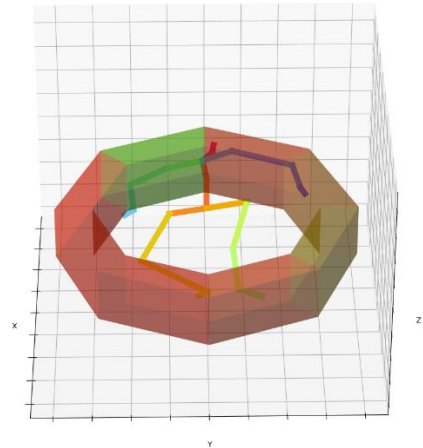

*Fig. 1 Proposed pressure measurement*

We propose a novel method to quantify the pressure on individual player with 3D body motion parameters considered. With this quantification, we manage to create the pressure matrix for each player which is further used to create a robust and easily interpretable "Player Pressure Map" (PPM) in the graph format to encapsulate this complex and multidimensional data. Each player on the possession team is treated as a node within a graph, with player interactions forming the edges. By incorporating contextual factors, including tracking data and players' movements, we embedded the information into the attacking players' nodes, creating the PPM. We train a Graph Neural Network (GNN) with PPMs as input to predict how likely the team is going to lose possession in the given game context. The probability of the attacking team losing possession in a specific game scene serves as a quantifiable measure of pressure from the opponent. This allows for the evaluation of player performance under varying levels of pressure, facilitating the recreation of specific pressure scenarios for more efficient training.

In conclusion, we propose a player pressure map to represent a given game scene, which lowers the dimension of raw data and still contains rich contextual information. Not only does it serve as an effective tool for visualizing and evaluating the pressure on the team and each individual, but it can also be utilized as a backbone for accessing players' performance. Overall, our model provides coaches and analysts with a deeper understanding of players' performance under pressure so that they make data-oriented tactical decisions.

# 1. Introduction

Over the last few years, high pressing in soccer has become a more and more popular defending tactic used by elite professional teams such as Manchester City F.C. and Liverpool F.C. The main idea of high pressing is to give the attacking opponent the pressure from the final third and force them to make mistakes under significant pressure. Since the pressure happens in the final third close to the opponent's goal, this tactic can enable the defending team to have a good chance of scoring in the transition if it is executed successfully [1]. Moreover, high pressing is used regularly not just by the strong teams but by smaller teams such as Gerona FC as well. Since effective high-pressing often doesn't need individuals to have extraordinary abilities, even small teams can get advantages by executing this tactic only if the players coordinate well in movement as a compact team [2].

Considering high pressing is commonly used nowadays in professional soccer games, the ability of the team and each player to handle the pressure and get away from the defenders pressing high on the pitch becomes more and more important. There have been multiple studies focusing on the analysis of pressure in soccer games. However, till now there is no commonly adopted approach to quantifying pressure. In research conducted by Taki et al., the pressure an individual player receives in the game was calculated according to the distances of this player to the opponents and to the ball [3]. The measurement of pressure with this method can be inaccurate when applied since it is based on the subjective definition of pressure without considering the other contextual factors deciding the pressure. Andrienko et al. improved the aforementioned pressure quantification method by considering the directions of the pressing target and the presser when calculating pressure[4]. However, the directions in this research are calculated only based on the ball movement without considering the player's body orientation.

In general, when quantifying the pressure around an individual player, there are two main limitations of current research. First, the measurement can be error-prone and inefficient when subjectively defining the factors deciding the pressure, such as the distance between the pressing player and the ball carrier. In addition, some of the important contextual factors are missed when quantifying the pressure that one player receives. For instance, the body orientation of a player plays an important role in deciding how much pressure he or she is facing.

When quantifying the pressure on the team, the best existing approach is integrating individual measurement into the team level quantification. However, this is not accurate since the team-level pressure should include other factors. For example, cutting the connections between the ball carrier and his/her teammates can also have a big impact on the level of pressure the attacking team receives while the aforementioned method fails to take these kinds of crucial factors into account.

Our research narrows the gap between current research about pressure in soccer and the answer to the main research question which is to accurately quantify the pressure the team and each individual player receive. We propose a novel pressure matrix to quantify the pressure on each individual player, by embedding rich contextual information including the probability that the defending team controls the specific areas and the body pose of the player with the ball. Then we propose Player Pressure Map (PPM), a graph integrating individual pressure and other crucial contextual information for the given game scene into 12 nodes representing attacking players and the ball and edges connecting all the nodes. For a specific game sequence, PPM is used to convert rich but complex contextual information into a unique sequence of embeddings, which makes it easier for the machine to learn meaningful patterns without sacrificing any important information extracted from the game. We train a possession outcome prediction model using Graph Neural Networks (GNN) with PPM sequences as input. The output of the trained model, representing the probability of the attacking team losing possession, is then used to quantify the pressure that the attacking team faces.

In general, our research has four major contributions:

1. We propose a novel method to quantify the pressure on each individual attacking player in a soccer game using tracking and event data and match footage.

2. We propose PPM, an insightful representation of team pressure in soccer.

3. We propose a novel possession outcome prediction model which can be used to quantify the team pressure effectively.

## 2. Methodology
### 2.1. Model architecture

To achieve the research objective, which is quantifying the pressure teams and individuals receive and evaluating their performances under different levels of pressure, we propose a machine learning framework. The overall architecture of the learning framework is shown in Fig. 2. For a given timestep, tracking and event data are used to generate a defensive pitch control matrix at first while the video footage of this game frame is used to generate the players' body motion parameters. The pressure matrix for each attacking player is then extracted from the pitch control matrix and fine-tuned with body motion parameters. There are two usages of the player pressure matrix. First, it is used to quantify the pressure on each player. Besides, player pressure matrices, together with player positions and velocities extracted from the tracking data, are used to generate the Player Pressure Map (PPM), a graph integrating contextual information for a given game scene. Next, we build a Graph Neural Network (GNN) [5]. To train a possession prediction model with PPMs as input. The output of trained GNN is used to quantify the team pressure.

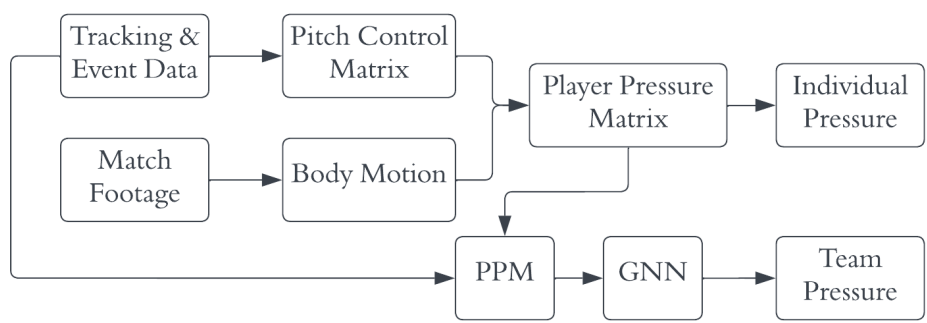

*Fig. 2 Proposed framework for measuring individual and team pressure*

## 2.2. Dataset

Tracking and event data and corresponding broadcast videos of 9 games in the Premier League are used in this research. Only active plays remain with penalties, set pieces and other occasions when the game is stopped removed from the datasets. Since the data of elite-level soccer games is often confidential, we only have access to 9 games at the moment of doing this research. Despite this limitation, we still have an ample amount of data to train our models and get ideal results.

## 2.3. Player pressure quantification

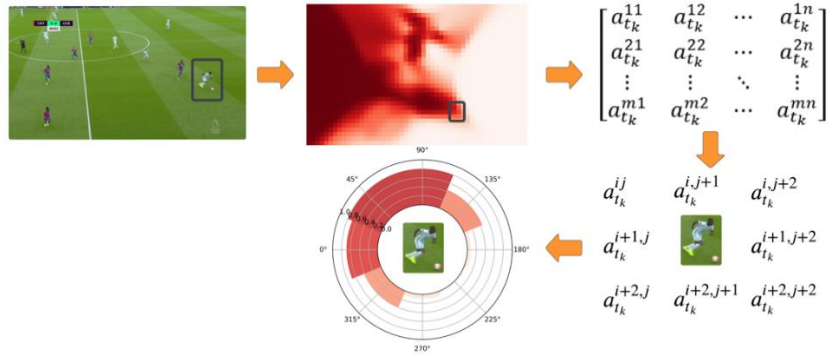

*Fig. 3 Individual spatial pressure measurement*

We first quantify the pressure each attacking player receives in a given game context. The method of quantification is shown in Fig. 3 With the pitch control function, at timestep $t_k$, for a given location (m, n) on the pitch, we calculate the probability of the defending team controlling the ball, $a_{t_k}^{mn}$, which is used to quantify the pressure on this location. To quantify the press on a specific player, we first locate his position on the pitch and use him as the center and 1 meter as the diameter to draw a pressure circle. From 8 directions around this player, the corresponding pressure values are calculated to create the vanilla pressure matrix for this player.

### 2.3.1 Player Orientation Extraction

Player body orientation is the direction the player is facing to. This has been proven to be an important contextual information during any in-game sequence [6]. While passing the ball, elite soccer players know how to use their body to block defenders and make a more efficient pass. In recent soccer research, models use calculated velocity from tracking data to represent the body orientation of the players [7]. However, [6] has shown the downside of using velocity directly.

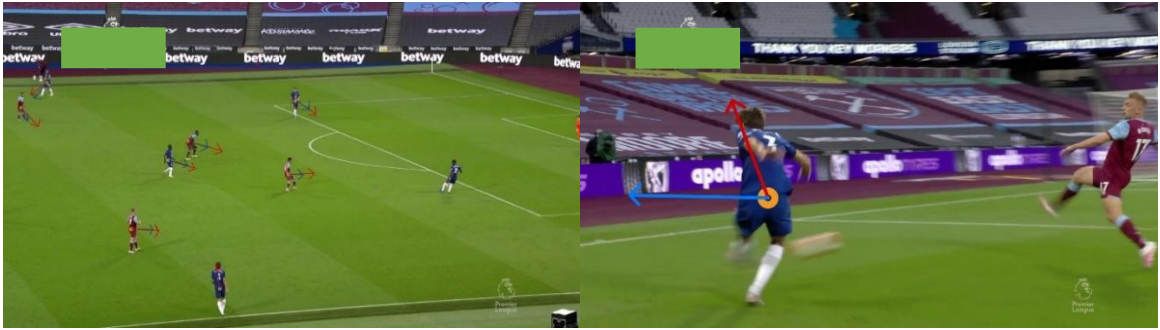

*Fig. 4 The body orientations of off-ball player and passing player*

As shown in Fig. 4, for most off-ball players, their velocity directions align with their body orientations. This is because when they are making an off-ball run, the main focus is to move themselves to an appropriate pitch location. However, for the on-ball player, they need to constantly adjust their body pose around the soccer. As shown in Fig. 4, the exact timestep when the on-ball player makes the pass, their body orientation changes drastically.

Usually, the player's 3D pose is required to extract their body orientation. However, the player 3D pose can only be captured by a costly depth-sensitive multi-camera system. To estimate player orientation directly from broadcast footage, [7] proposed a computer vision model to detect the 2D poses of the players and the homography transformation that projects the soccer pitch from broadcast footage to a 2D template. The downside of this approach is that it is rather difficult to reimplement since both the 2D pose detection model and the pitch transformation detection model requires a huge amount of training data. Annotation for training these two models is very time consuming. Even with the annotated data, the generalization results on differently distributed inference data is not accurate enough for usage in industry [8]. A more efficient and accurate approach to extract the body orientation of the passer could be very beneficial for soccer analytics.

To estimate the passer's body orientation at the exact moment of passing. We followed a top-down approach to estimate player 3D pose using broadcast footage as input, meaning that we first detect a bounding box for each player, and then for each detected player we can perform a 3D pose estimation model. Out of all detected players, the tricky task is to determine which one is the passer. During our experiments, we tried two other computer vison approaches. Firstly, a soccer detection model was used to also detect the soccer pixel coordinates, and the player whose bounding box is the closest to the ball is determined to be the passer. In the broadcast footage, the soccer is constantly moving and usually above the player's body level, which leads to the accuracy of this approach unusable for our analysis. Secondly, we trained a homography estimation model that enables us to project detected player onto a 2D soccer pitch. By matching the trajectory of projected player to the actual tracking and event data, we can detect which player is the passer. The accuracy of this approach is highly dependent on the accuracy of the homography estimation model. Since the soccer broadcast footage contains a large amount of camera movement, the accuracy of estimating the homography becomes unreliable. We found that the most reliable way of determining the passer from all the detected players is to assign an identification to all the detections and a human expert is required to annotate the detection with the identification of the passer.

To achieve higher annotation efficiency and 3D pose estimation accuracy, besides the single frame where the passing happened, we also detected players in the frames within 0.25 second. By using a tracking model, we can track the players with the same identification across the frames [9]. While we only annotated the passing frame, the tracking model enabled us to determine the passer in all the frames. After detecting and determining the passer in the whole passing sequence, we can estimate a sequence of 3D body pose of the passer. This is useful for smoothing out the 3D body motion of the passer and then improve the 3D pose estimation accuracy at the passing frame when there is body occlusion in the broadcast footage.

All the 3D pose estimation models estimate the 3D poses in the coordinate system with the camera focal at the origin and the axis direction determined by the camera extrinsic [10]. Therefore, all of them struggle to localize the players in the world coordinate, which usually requires a transformation function to rotate, transform and scale the estimated 3D pose. However, based on our observations, directly transforming the 3D pose estimation to the soccer pitch based on the tracking data without rotating and scaling does not impact the accuracy of estimated body orientation.

### 2.3.2. Individual Player Pressure Matrix Fine-tuning

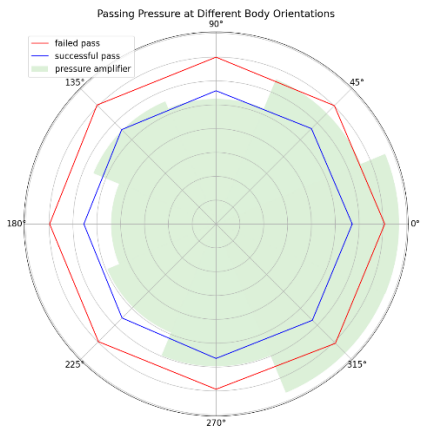

*Fig. 5 Passing pressure in relative to body orientation*

With the body motion of the passing player detected from the game footage, we get the orientation of his body. Following the proposed framework, the passer body orientation was extracted for all the passes in six soccer games. As shown in Fig. 5, the first observed pattern is that the average pressure from all directions during a failed pass is higher than the successful pass. This further indicates that the proposed pressure matrix is able to indicate the spatial pressure level for individual players.

To finetune the pressure matrix for the passer, we are trying to answer one critical question: the pressure from which direction is more impactful for the pass outcome. We also observed that the passer experiences higher pressure in the front and to the right of their body orientation. This is also agreed by the fundamental fact of soccer, the defender is usually trying to apply pressure opposite to the body orientation of the possessing player. Given the body orientation of the passer, to amplify the importance of the pressure from key directions, we calculated a pressure amplifier based on observed passing data as shown in Fig.5.

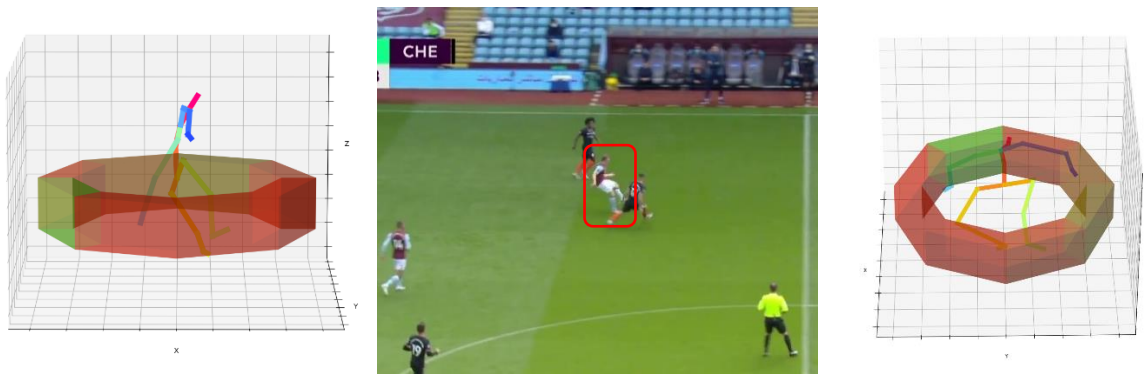

*Fig. 6 finetuned pressure matrix viewed in $90°$ and $180°$ (opposite to attacking direction).*

The orientation is then used to compute a pressure amplifier, which is then utilized to fine-tune the pressure matrix of the passer based on their body orientation. The fine-tuned pressure matrix shown in Fig.6 can reflect the pressure on this player more accurately by considering the 3D contextual factors.

## 2.4. Player Pressure Map (PPM)

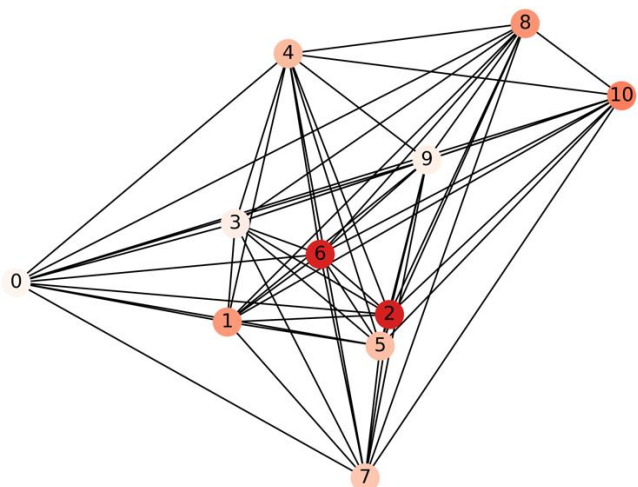

*Fig. 7 Player pressure map (PPM)*

With the above pressure quantification method, for a game scene, each attacking player has a pressure vector to quantify the pressure he receives at the given moment. For a specific player, the pressure vector, together with his position, is embedded into the node representing him. To connect all the 11 nodes representing the attacking players, we use edges that include the distances and angles between each two players. Compared to the graph representations in the previous studies on soccer analytics [11, 12], our PPM has a lower dimension with only attacking players included. Besides, defending players' contextual information is embedded into the graph nodes through pressure vectors, which enables the graph to avoid losing important features when used to represent the given game scene. Moreover, we extract important body motion features from match footage and integrate them into the graph, which makes the PPM include 3D contextual factors of a specific game context.8

## 2.5. Team pressure quantification

To quantify the pressure of the attacking team received, we train Possession Outcome Prediction (POP) model, which is used to predict whether the attacking team is going to lose control of the ball in the coming seconds at a certain moment. And the predicted possession loss (pop) probability, values from 0 to 1, is used to represent the received team pressure. In the case of the pop value equals to 0.64, that means the attacking team has a 64% chance to remain the possession in the next 4 seconds, while 36% possibility to lose. The higher chance a team is going to lose control means the higher pressure they receive from the opponents in this research, vice versa.

We firstly filter all the possessions longer than 5 seconds from tracking dataset, and every two-second possession is converted into a sequence of 50 PPMs. The PPMs are then fed into a Graph Neural Network to predict the outcome of this possession after 4 seconds. If the team loses possession, then the outcome will be labelled as 0 and if the team keeps possession, the label will be 1. Any possession moment or frame whether the attacking team is passing or dribbling or doing any other actions can be fed into the trained model, the outcome will be a team-level pressure metric based on the pop value.

The architecture of GNN is listed as follows. The model contains three Graph Convolution Layers to extract the features from the input dataset, including tracking and event data, for instance, the x and y locations of each player, the velocity of each player and the distance between each player. A ReLu activation layer and a Global Mean Pooling layer are applied to each Graph Convolution Layer [13]. The dropout rate of the whole model is 0.5 and eventually a fully connected layer is connected to generate the 2 classes prediction outcome.

## 3. Results

|  | Tracking | 2D PPM | 3D PPM |
|---|---|---|---|
| Prediction Accuracy | 55.8% | 75.2% | 78.7% |

Table 1. Model Performance Metrics

To evaluate the model performance, we compare the prediction accuracy from two benchmarks and our proposed one. The test dataset is one independent Premier League match from the same season, which has never been used to train any component of this model. More than 750 possessions have been generated as the test dataset. Table 1 shows the prediction accuracy for all three models. Prediction accuracy is defined as the success rate of the possession outcome between the estimated and the truth. The first baseline model is trained only based on the tracking dataset, which is the normal way of previous research, and the prediction accuracy is 55.8%. The second benchmark model is trained based on 2D PPM; the accuracy rises to 75.2% with a significant increase of around 20%. The improvement proves that by only relying on the tracking dataset, the previous method does not contain the complicated context of the ever-changing football pitch. While the proposed model by us is a step further, the 3D PPM model performs around 3% better than the second benchmark model. The strength of our model is largely contributed by including more 3D features from players and pitch. The 3D features from broadcasting videos are not all included and how to represent the high dimension 3D features within and between individuals effectively still needs to be explored, but we do believe that our method and model prove the importance of 3D features and their impact on the pressure metrics, both of which can improve the potential of the tracking dataset in the area of sports performance analysis.

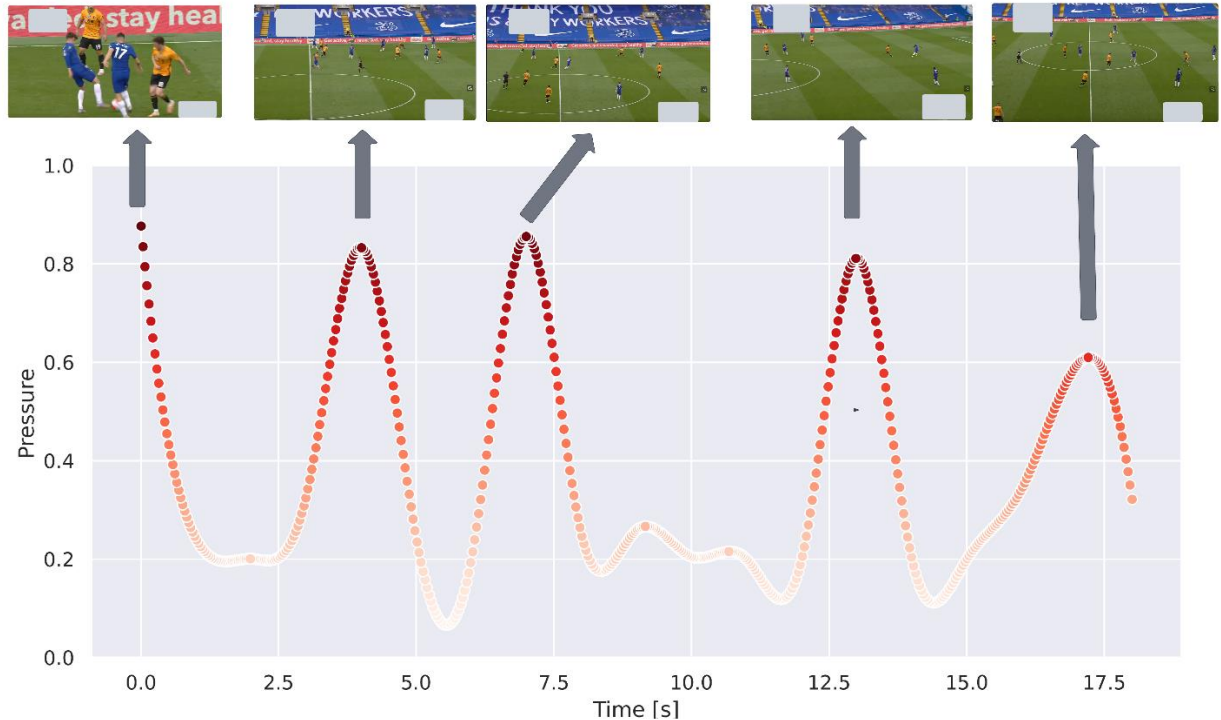

*Fig. 8 Team pressure quantification example*

In the case of this possession shown in Fig.8, the blue team tackles the ball back from the opponent in the right midfield, since the ball carrier is surrounded by multiple opponent players, the risk of losing possession right now is relatively high [14]. The ball carrier decides to pass back to the center defender who dropped back, as he makes the backward pass, the possession rate rockets up rapidly. The center defender who received the ball chose to switch the attacking direction. During the switching process, the defending team performs patiently thus the pressure metric remains at a low level until the blue team finishes their switching and attempts to attack from another side. That attempt of passing the ball to the blue team midfielder triggers another round of pressure as we can see the pressure metrics increase rapidly several times. The whole high-line pressure ends until the blue team goalkeeper is forced to kick the long ball and give the possession out under the pressure from the whole opponent team. Now, besides the broadcasting video, our proposed method provides a direct quantitative metric to measure the intensity of the pressure at team level.

## 4. Discussion

### 4.1 Player performance under different levels of individual pressure

With our method of quantifying pressure on individuals, we calculate the passing accuracy of all players under different levels of pressure on the whole dataset. As shown in Fig. 9. Among all positions, midfielders are usually under the highest pressure. Moreover, compared to the defenders and attackers, midfielders can perform more consistently under different levels of pressure. Attackers have the lowers passing accuracy under all levels of pressure, which is reasonable since they are trapped in compact defensive squad more often than players from the other positions. Besides, we select two players who both make more than 300 passes to compare their performance taking pressure into consideration. If we only look at the general passing accuracy of two players, the numbers are close to each other, around 80 percent. However, with our pressure metric, we can clearly find the difference between these two players' abilities. Player 184341 can handle the pressure better compared to player 225796 with more consistent passing accuracy under different levels of pressure while player 226796 has a very bad passing success rate when the pressure on him is level 3. In real life, player 184341 is one of the best attacking midfielders in Premier League of that season while player 225796 is an average stay back center defensive midfielder, which matches the results of our model. Overall, our method can make the player performance evaluation be more consistent and accurate with pressure considered.

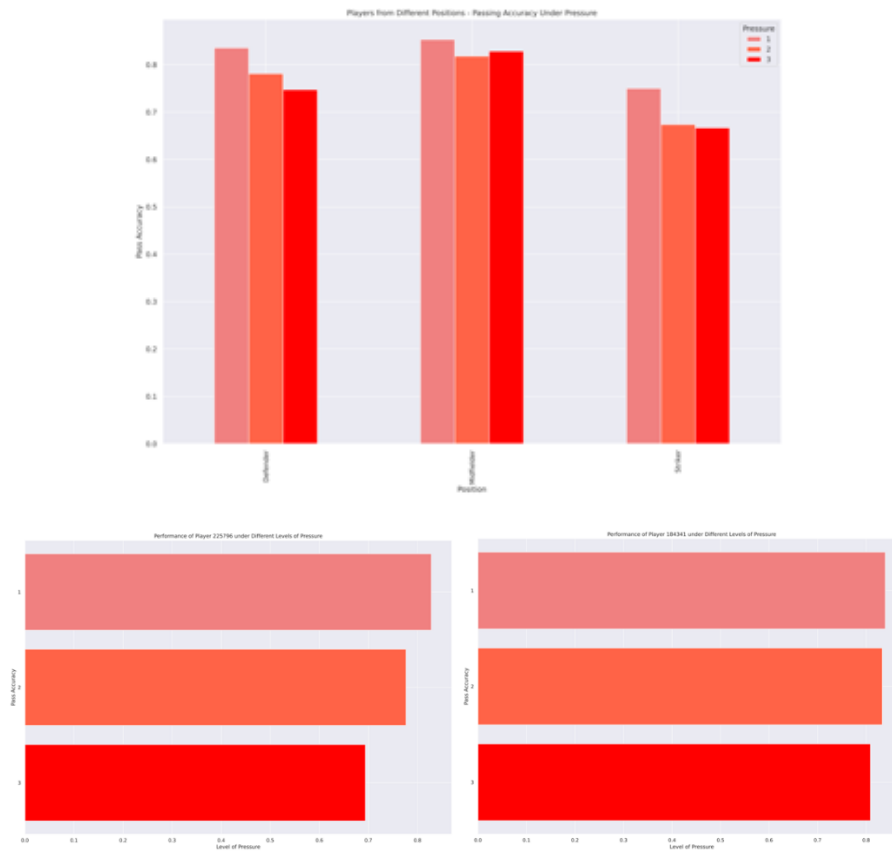

*Fig. 9 Players' passing accuracies under different pressure levels*

*(level 1: pressure <= 1/3; level 2: 1/3 < pressure <= 2/3; level 3: pressure >2/3)*

## 4.2 Player performance under different levels of team pressure

With our trained possession outcome prediction model, we can quantify the team pressure for a given match frame using the model output, the probability of losing the possession. With this team pressure quantification method, we analyze the players' performance under different levels of team pressure on the dataset from the testing match.

We quantify the passing and dribbling events within the game. For a specific event, to evaluate the player's performance, we calculate the team pressure at starting point and ending point and use the amount of decrease between two pressure values to determine how well the player performs in terms of contributing to relieving the pressure on his team. As shown in Fig 10. Player 41328 is the most effective dribbler of this game. He can decrease the pressure on his team by around 0.4 per dribble. Moreover, player 38533 is the most efficient passer of this game by relieving the pressure on the team through incisive passes. Considering the overall performance in passing and dribbling, player 41328 outperforms the rest in terms of diminishing the pressure when his team attacks. Player 41328 has the second highest match rating in that game, which matches the evaluation enabled by our model. In general, with our method, coaching staff can evaluate players' contributions to confronting opponent's pressure. Moreover, for each event within the game, coaching staff can directly locate the ineffective dribbles and passes and make data-oriented adjustments.

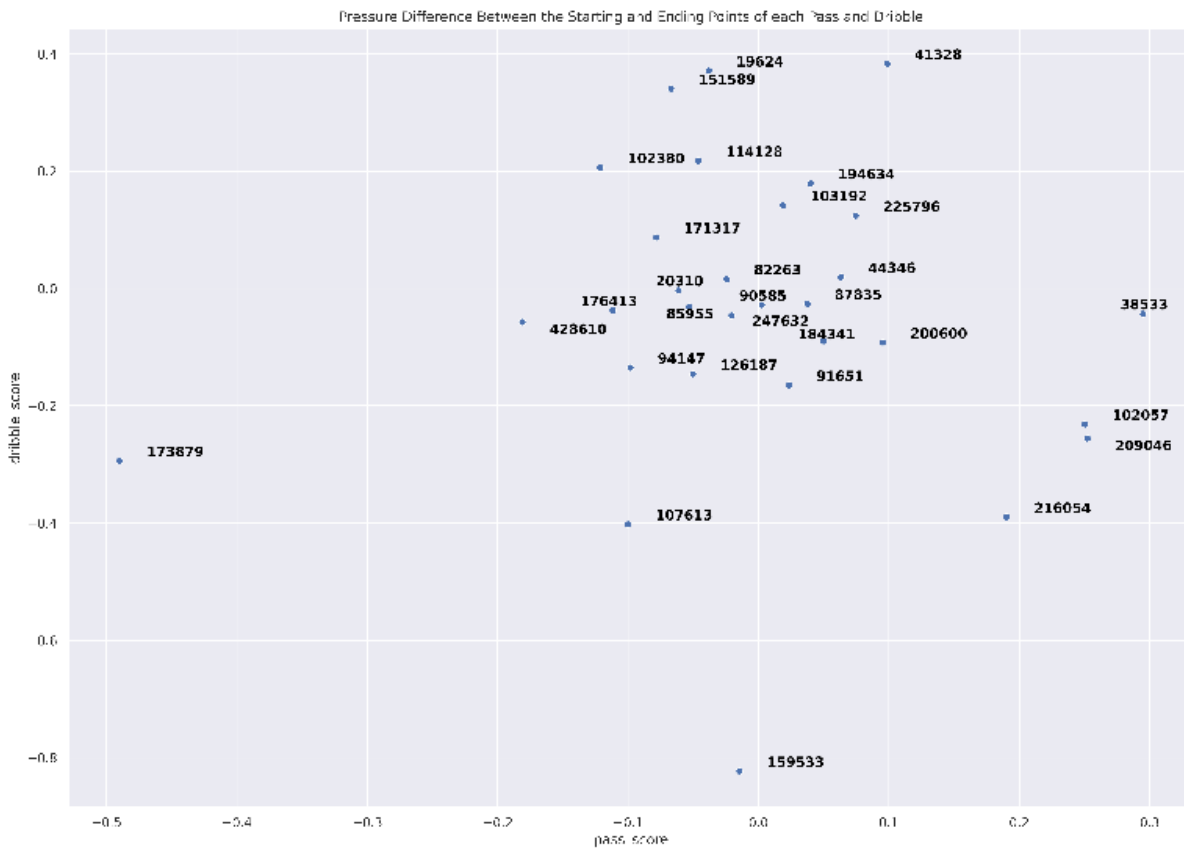

*Fig. 10 Pressure difference between starting and ending points of each pass and dribble*

# 5. Conclusion

The strength of high pressing tactics in soccer speak for itself, the challenge here is how to evaluate the contribution and impact of the pressure on both individual and team level, the player performance then could be ranked and compared to reveal their capabilities of handling stress and decision making. In this study, we trained our model with tracking and event datasets and corresponding matches footages collected from 9 Premier League matches in 2019/20 season and quantify the pressure on the attacking team and each individual during possession and evaluated player performance in various contexts. The model results show that our metrics can detect the pressure within the game precisely. The ideal results can be attributed to the novel embedding pf contextual information with 3D body motions considered and the utilization of graph neural network. Moreover, our pressure quantification method has proved to be effective in evaluating players' performance. When judging players' passing and dribbling abilities, taking pressure into consideration makes the evaluation more robust and accurate, which can strengthen the effectiveness of coaching staff's decision-making in after-game analysis and help scouts understand the players' strengths and weaknesses better.

In future work, we can utilize the temporal series machine learning models such as Long short-term memory networks to expand input contextual information from one frame to multiple frames, which is expected to improve the model performance.

# References

[1] Low, Benedict, et al. "The porous high-press? An experimental approach investigating tactical behaviours from two pressing strategies in football." *Journal of Sports Sciences* 39.19 (2021): 2199-2210.
[2] Andrienko, Gennady, et al. "Exploring pressure in football." *Proceedings of the 2018 International Conference on Advanced Visual Interfaces*. 2018.
[3] Taki, Tsuyoshi, and Jun-ichi Hasegawa. "Dominant region: a basic feature for group motion analysis and its application to teamwork evaluation in soccer games." *Videometrics VI*. Vol. 3641. SPIE, 1998.
[4] Andrienko, Gennady, et al. "Visual analysis of pressure in football." *Data Mining and Knowledge Discovery* 31 (2017): 1793-1839.
[5] Zhou, Jie, et al. "Graph neural networks: A review of methods and applications." *AI open* 1 (2020): 57-81.
[6] Arbués-Sangüesa, Adrià, et al. "Learning football body-orientation as a matter of classification." *arXiv preprint arXiv:2106.00359* (2021).
[7] Arbués-Sangüesa, Adrià, et al. "Head, shoulders, hip and ball... hip and ball! using pose data to leverage football player orientation." *2nd Barça Sports Analytics Summit* (2019).
[8] Bhattad, Anand, et al. "View generalization for single image textured 3d models." *Proceedings of the IEEE/CVF Conference on Computer Vision and Pattern Recognition*. 2021.
[9] Aharon, Nir, Roy Orfaig, and Ben-Zion Bobrovsky. "BoT-SORT: Robust associations multi-pedestrian tracking." *arXiv preprint arXiv:2206.14651* (2022).
[10] Chen, Ching-Hang, and Deva Ramanan. "3d human pose estimation= 2d pose estimation+ matching." *Proceedings of the IEEE conference on computer vision and pattern recognition*. 2017.
[11] Xenopoulos, Peter, and Claudio Silva. "Graph neural networks to predict sports outcomes." *2021 IEEE International Conference on Big Data (Big Data)*. IEEE, 2021.

[12] Zheng, Nan, Meng Sun, and Ye Yang. "Visual Analysis of College Sports Performance Based on Multimodal Knowledge Graph Optimization Neural Network." *Computational Intelligence and Neuroscience* 2022 (2022).
[13] Zhang, Si, et al. "Graph convolutional networks: a comprehensive review." *Computational Social Networks* 6.1 (2019): 1-23.
[14] Low, Benedict, et al. "The porous high-press? An experimental approach investigating tactical behaviours from two pressing strategies in football." *Journal of Sports Sciences* 39.19 (2021): 2199-2210.